\documentclass[runningheads]{llncs}
\usepackage{graphicx}
\usepackage{subcaption}
\usepackage{booktabs}
\usepackage{amsmath}
\usepackage{amssymb}
\usepackage{array}
\usepackage{placeins}
\usepackage[colorlinks]{hyperref}

\usepackage{pifont}
\newcommand{\cmark}{\ding{51}}%
\newcommand{\xmark}{\ding{55}}%

\begin{document}
\title{Sanity Checks for Explanation Uncertainty}
\author{Matias Valdenegro-Toro\inst{2}\orcidID{0000-0001-5793-9498} \and Mihir Mulye\inst{1}\orcidID{0009-0001-8043-8752}}
\authorrunning{Valdenegro-Toro \& Mulye}
\institute{Department of Artificial Intelligence, University of Groningen, 
       9747AG Groningen, The Netherlands \quad \email{m.a.valdenegro.toro@rug.nl} \and
Bonn-Rhein-Sieg University of Applied Sciences, 53757 Sankt Augustin, Germany \newline
\email{mihirmulye95@gmail.com}}
\maketitle              %
\begin{abstract}

Explanations for machine learning models can be hard to interpret or be wrong. Combining an explanation method with an uncertainty estimation method produces explanation uncertainty. Evaluating explanation uncertainty is difficult. In this paper we propose sanity checks for uncertainty explanation methods, where a weight and data randomization tests are defined for explanations with uncertainty, allowing for quick tests to combinations of uncertainty and explanation methods. We experimentally show the validity and effectiveness of these tests on the CIFAR10 and California Housing datasets, noting that Ensembles seem to consistently pass both tests with Guided Backpropagation, Integrated Gradients, and LIME explanations.

\end{abstract}
\section{Introduction}
Neural networks are used for many tasks, producing impressive performance, but still they are black-box models that are very difficult to interpret for humans. There is increasing interest on obtaining explanations from neural networks, due to their increasing use and for legal and ethical requirements \cite{atkinson2020explanation}.

There are many ways to attempt to obtain interpretations for a neural network prediction, including gradient-based saliency methods \cite{borys2023explainable}, interpretable models, and gray-box models \cite{rudin2019stop}. Evaluating explanations is particularly difficult since they are ambiguous and there is no ground truth explanation \cite{samek2016evaluating}.

\begin{figure}[t]
    \centering
    \begin{subfigure}{0.32\textwidth}
        \includegraphics[height=2.95cm]{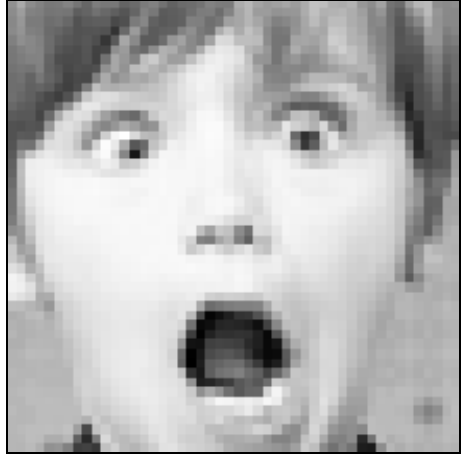}
        \caption{Input\\Image}
    \end{subfigure}
    \begin{subfigure}{0.32\textwidth}
        \includegraphics[height=3cm]{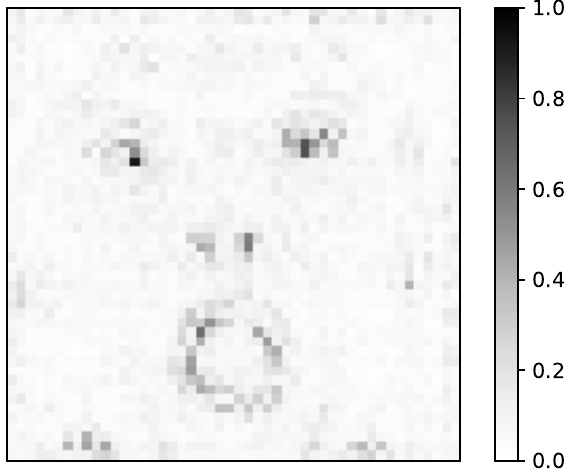}
        \caption{Explanation\\Mean}
    \end{subfigure}
    \begin{subfigure}{0.32\textwidth}
        \includegraphics[height=3cm]{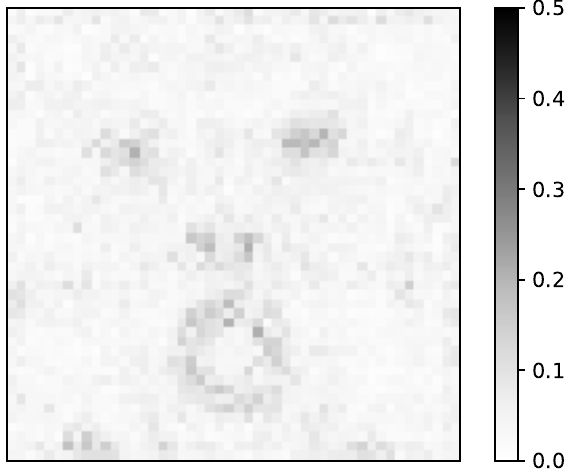}
        \caption{Explanation\\Uncertainty}
    \end{subfigure}
    \caption{Example explanation with uncertainty on a FER dataset image using Guided Backpropagation and Ensembles.}
    \label{example_expl}
\end{figure}

An explanation is often assumed to be correct, but as any prediction, explanations can also be misleading or incorrect, which motivates the need to quantify explanation uncertainty, as show in Figure \ref{example_expl}.

There is previous work about producing explanations with uncertainty, like Slack et al. \cite{slack2021reliable} that use a Bayesian framework to model explanation uncertainty in LIME and KernelSHAP, Bykov et al. \cite{bykov2021explaining} that combines Bayesian Neural Networks and Layer-wise Relevance Propagation to produce explanation uncertainty, and Zhang et al. \cite{zhang2019should} that find sources of uncertainty in LIME: sampling randomness, sampling proximity, and explanation variation across data points.

Uncertainty estimation methods are often approximations, to a Bayesian Neural Network (BNN) \cite{lampinen2001bayesian}, or using non-Bayesian methods like Ensembles \cite{lakshminarayanan2016simple}. Quantifying the quality of explanation uncertainty is an open problem, as standard metrics for explanations cannot be directly applied, motivating the need for alternative testing methods.

In this paper, we explore the idea of extending sanity checks \cite{adebayo2018sanity} to evaluate the quality of explanation uncertainty in gradient-based explanation methods. We combine several uncertainty estimation method (MC-Dropout, MC-DropConnect, Ensembles, Flipout) with saliency explanations like Guided Backpropagation, Integrated Gradients, and LIME, and propose sanity checks based on data and weight randomization, to validate if explanation uncertainty does behave according to some basic expectations: that when the model is manipulated to lose information (randomization), this loss of information is reflected in the explanation uncertainty, by increasing explanation uncertainty as randomization progresses.

This work aims to increase understanding on how explanations behave and if explanation uncertainty follows some basic rules, as uncertainty in an explanation should be somehow proportional to correctness, and there is no direct way to evaluate explanation correctness. Sanity checks can be used as basic test when developing new explanation methods with uncertainty estimation.

The contributions of this work are: we propose weight and data randomization tests for explanations with uncertainty, setting some basic expectations as explanation uncertainty should increase as tests progress, and we provide an initial evaluation of several uncertainty estimation methods and saliency explanation methods on CIFAR10 and California Housing datasets.

\section{Sanity Checks for Uncertain Explanations}

\subsection{Explanation Uncertainty}

This section introduces the basic notation and concept about explanations with uncertainty.

Given a trained machine learning model $f_\theta$ with parameters $\theta$, we assume that the model has integrated some method for uncertainty quantification, such as MC-Dropout, Ensembling, Variational Inference, etc. Quantifying uncertainty usually implies multiple stochastic forward passes (MC-Dropout, Variational Inference), or multiple models (Ensembles):
\begin{equation}
    \mu(x) = T^{-1} \sum_i f_\theta(x)
\end{equation}
\begin{equation}
    \sigma^2(x) = T^{-1} \sum_i (f_\theta(x) - \mu(x))^2
\end{equation}
With $T$ the number of stochastic forward passes or ensemble members. To obtain a gradient-based saliency explanation $\text{expl}(x)$, it is often a function of model gradients and predictions $\hat{y} = f_\theta(x)$:
\begin{equation}
    \text{expl}(x) = F\left(\hat{y}, \frac{\partial f_\theta(x)}{\partial x}\right)
\end{equation}
To produce explanations with uncertainty, we combine the two previous formulations, noting that with uncertainty model, the gradient is also stochastic, and across multiple forward passes or ensemble members, a explanation mean $\text{expl}_\mu(x)$ and standard deviation $\text{expl}_\sigma(x)$ can be computed:
\begin{equation}
    \text{expl}_\mu(x) = T^{-1} \sum_i \text{expl}(x) = T^{-1} \sum_i F\left(\hat{y}, \frac{\partial f_\theta(x)}{\partial x}\right)
\end{equation}
\begin{equation}
   \text{expl}_\sigma(x) = T^{-1} \sum (\text{expl}(x) - \text{expl}_\mu(x))^2
\end{equation}
The standard deviation $\text{expl}_\sigma(x)$ can be interpreted as multiple forward passes will produce different predictions, together with different explanations. If the explanations are all very different, this will produce a high standard deviation, and if explanations are similar, then standard deviation will be low. Then $\text{expl}_\sigma(x)$ can be used to assess if the combination of model, uncertainty quantification method, and gradient explanation method, produce explanations that are robust to multiple forward passes (low $\text{expl}_\sigma(x)$, or the explanation is low quality and not clear to the user (high $\text{expl}_\sigma(x)$).

A Bayesian interpretation is that a posterior predictive distribution for the explanation can be computed through a gradient explanation, and in this paper we use monte carlo approximations.

Explanations computed with the formulation above, combining uncertainty quantification methods with explanation ones, are denoted as explanation uncertainty. Explanations without uncertainty quantification can be named as point-wise explanations, in consistency with the uncertainty quantification literature.

Note that explanation uncertainty is explicitly about modelling uncertainty in the explanation, while uncertainty explanation is a different concept, where the aim is to explain the uncertainty of a prediction.

For visualization purposes, it is possible to integrate the mean and standard deviation of an explanation into a single explanation, using the coefficient of variation $\text{CV}(x)$:
\begin{equation}
    \text{CV}(x) = \frac{\text{expl}_\sigma(x)}{\text{expl}_\mu(x)}
\end{equation}

\subsection{Sanity Checks}

This section introduces sanity checks for explanation uncertainty.

Sanity checks for saliency explanations were originally introduced by Adebayo et al. \cite{adebayo2018sanity}, and they were defined for point-wise explanations. In this section we define how these sanity checks can be expanded to consider explanation uncertainty, and how they can help validate it. There are two basic sanity checks:

\textbf{Weight Randomization}. Model weights are randomized progressively, layer by layer, starting by the layers closest to the input. The concept is that by destroying information stored in weights, through randomization, the explanation should also be close to random, and explanation uncertainty $\text{expl}_\sigma(x)$ should increase, and keep increasing as more layers are randomized.

Explanation uncertainty  $\text{expl}_\sigma(x)$ should be a measure of correctness for a specific explanation, so naturally a correct explanation should have low uncertainty, and a wrong explanation should have high uncertainty. By destroying information in the model, the explanation uncertainty $\text{expl}_\sigma(x)$ should reflect this loss of information in the model, in a way it is a form of epistemic uncertainty.

We propose to evaluate the weight randomization sanity check by training a model on a given dataset, create explanations with uncertainty on the test set, and randomize weights progressively, layer by layer, and compare the explanation uncertainty  $\text{expl}_\sigma(x)$ on the test set as layers are randomized, and we expect to see increasing $\text{expl}_\sigma(x)$ compared to the explanation uncertainty on the trained model on the test set.

\textbf{Data Randomization}. Training set labels are randomized, and a model is trained on randomized labels, which should lead the model to overfit and not generalize to a validation or test set. Explanations with uncertainty are made, and explanation uncertainty  $\text{expl}_\sigma(x)$ should be higher for the model trained on random labels than for the model trained on the real labels.

The concept of label randomization is to train an improper model, where the learned relationship between inputs and outputs does not generalize outside the training set. Explanations in this case are expected to not be informative, and this should be reflected in the explanation uncertainty $\text{expl}_\sigma(x)$.

Both sanity checks for explanation uncertainty allow for basic evaluation of explanation uncertainty methods, as there are no other methods to specifically evaluate explanation uncertainty. These checks test the specific combination of uncertainty estimation and explanation methods, as both interact, as will be shown in our results.

\begin{table}[t]
    \begin{tabular}{p{6.3cm}p{0.2cm}p{5.5cm}}
        \toprule
        \textbf{Description} & & \textbf{Expected Outcome}\\
        \midrule
        \multicolumn{2}{l}{\textbf{Weight Randomization Test}} \\
        Randomize weights of the model progressively, destroying task information stored on weights & & Explanation uncertainty increases as weights are progressively randomized.\\
        \midrule
        \multicolumn{2}{l}{\textbf{Data Randomization Test}} \\
        Train a model on randomized labels, which produces a model unable to generalize & & Explanation uncertainty is higher compared to a model trained on proper labels.\\
        \bottomrule
    \end{tabular}
    \caption{Summary of sanity checks for explanation uncertainty and proposed expectations when performing these tests. This small guide can be used to decide if a XAI + UQ method passes or fails a test.}
\end{table}

\section{Experiments}

This section describes our evaluation of several combination of saliency explanation methods and uncertainty quantification methods. We experimentally assess if these combinations pass each test or not. The aim of our evaluation is not to provide a definite answer if these methods pass our sanity checks or not, but to validate that the proposed sanity checks do work and provide a way to test and evaluate explanation uncertainty methods.

We present results on CIFAR10 Image Classification, and California Housing price regression. First we describe the saliency methods we evaluate, and then the uncertainty estimation methods, then present our actual experimental results, and then close with some discussion and summary.

\subsection{Saliency Methods}

This section describes the saliency methods we used to evaluate explanation uncertainty.

\textbf{Guided Backpropagation}. Gradient-based saliency methods produce explanations by computing $\frac{\partial \hat{y}}{\partial x}$, the gradient of network output $\hat{y}$ (for a particular output neuron) with respect to input $x$, which measures impact of the network output as the input changes, which is a form of explanation.

GBP \cite{springenberg2014striving} modifies the gradient of the ReLU activation to reduce explanation noise and enhance human interpretability. The activation $f$ after applying ReLU activation $\textit{relu}$ at layer $l$ is:
\begin{equation}
    \label{eq:activation}
    f_{i}^{l+1}=\operatorname{\textit{relu}}\left(f_{i}^{l}\right)=\max \left(f_{i}^{l}, 0\right)
\end{equation}

Where $i$ is the feature dimension. GBP handles backpropagation through ReLU non-linearity by combining vanilla backpropagation and DeconvNets as: 

\begin{equation}
    \label{eq:gbp}
    R_{i}^{l}=\left(f_{i}^{l}>0\right) \cdot\left(R_{i}^{l+1}>0\right) \cdot R_{i}^{l+1}
\end{equation}

The reconstructed image $R$ at any layer $l$ is generated by the positive forward pass activations $f_i^l$ and the positive error signal activations $R_i^{l+1}$.  This aids in guiding the gradient by both positive input and positive error signals. The negative gradient flow is prevented in GBP, providing more importance to neurons increasing the activation at higher neurons. 

\textbf{Integrated Gradients}. (IG) \cite{sundararajan2017axiomatic} aims to improve explanations by setting axioms and expectations, and designing a method that fulfills them. IG performs an integral along a linear path from a baseline input $x'$ to given input $x$ of the local gradient. IG computation for feature $i$ is:
\begin{equation}
    \text {IG}_{i}(x, F) =\left(x_{i}-x_{i}^{\prime}\right) \times \int_{\alpha=0}^{1} \frac{\partial F\left(x^{\prime}+\alpha \times\left(x-x^{\prime}\right)\right)}{\partial x_{i}} d \alpha
    \label{equ:ig}
\end{equation}
$\alpha$ is the interpolation constant for feature perturbation the straight path between baseline and input. 
$F(x)$ is the model function mapping features to predictions. 
The solution is obtained using numerical approximation because calculating definite integral for Equation \ref{equ:ig} is difficult.
\begin{equation}
    \text {IG}_{i}(x, F) =\left(x_{i}-x_{i}^{\prime}\right) \times m^{-1} \sum_{k=1}^{m} \frac{\partial F\left(x^{\prime}+\frac{k}{m} \times\left(x-x^{\prime}\right)\right)}{\partial x_{i}}
    \label{equ:ig}
\end{equation}
\textbf{LIME.} This method \cite{ribeiro2016should}	produces global explanations using a local method (linear regression), by first sampling around the desired input $x$ to be explained and then fitting a surrogate method that can produce a local explanation. The core idea of LIME is to locally approximate the decision boundary of the model and then make a local explanation from this approximation.

\subsection{Uncertainty Quantification Methods}

This section describes the uncertainty quantification methods we used to evaluate explanation uncertainty. We first cover three stochastic methods, where the model becomes stochastic and each forward pass produces a sample of a predictive distribution, estimating model uncertainty.

\textbf{Dropout}. Monte Carlo Dropout \cite{gal2016dropout} is a well known method for uncertainty estimation, corresponding to enabling Dropout at inference time, the model becoming stochastic, and each forward pass producing one sample of the predictive posterior distribution. We use dropout layers with $p = 0.5$.

\textbf{DropConnect}. Monte Carlo DropConnect \cite{mobiny2021dropconnect} is similar to MC-Dropout, but weights are dropped (set to zero) instead of activations, and this behavior is enabled at inference time, with forward passes also producing a sample of the predictive posterior distribution.

\textbf{Flipout}. This method is a Bayesian Neural Network \cite{blundell2015weight}, approximating weights $W$ with a Gaussian distribution $q_\theta(W)$, using a loss that approximates the Evidence Lower Bound (ELBO):

\begin{equation}
    L(\theta) = \text{KL}(\mathbb{P}(W), q_\theta(W)) - \mathbb{E}_{w \sim q_\theta(W)}[\log \mathbb{P}(y \, | \, x, w)]
\end{equation}

These loss terms are the KL-divergence between the prior $\mathbb{P}(W)$ and the approximate weight posterior distribution $q_\theta(W)$, plus the negative log-likelihood $-\log \mathbb{P}(y \, | \, x, w)$ computed as an expectation by sampling weights from the approximate weight posterior $w \sim q_\theta(W)$. The negative log-likelihood is the loss function for the task, mean squared error for regression, and cross-entropy for classification.

In particular Flipout \cite{wen2018flipout} stabilizes training by allowing different sampled weights to be used in a batch of predictions, speeding up training and improving predictive performance.

We also use one deterministic method, ensembling:

\textbf{Ensembles}. Ensembling \cite{lakshminarayanan2016simple} trains $T$ copies of the same model architecture, on the same dataset, and then combines their predictions as shown below. Differences between models come from random weight initialization. Ensembles are often one of the best performing uncertainty estimation methods, and are very easy to implement.

As all uncertainty quantification methods we use produce samples or use multiple models, those samples/predictions have to be combined to estimate model uncertainty. For classification, the predicted probability vectors are averaged, and for regression, the mean $\mu(x)$ and variance $\sigma^2(x)$ across samples/predictions is computed:

\begin{equation}
    \mu(x) = T^{-1} \sum f(x) \qquad \qquad \sigma^2(x) = T^{-1} \sum (f(x) - \mu(x))^2
\end{equation}

Where $T$ is the number of samples/models. The standard deviation $\sigma(x)$ is a measure of uncertainty in this case.

\subsection{CIFAR10 Classification}

In this section we perform experiments using the well known CIFAR 10 dataset. We train a small CNN on this dataset using MC-Dropout for uncertainty estimation, and perform sanity checks on Guided Backpropagation (GBP) and Integrated Gradients (IG).

As visually validating image explanations is not easy, in order to compare explanations produced by a modified model (data and weight randomization tests) with the explanations made by the original (unrandomized) model, we use the Structural Similarity Index (SSIM).

\begin{figure}[t]
    \centering
    \begin{subfigure}[b]{0.475\textwidth}
        \centering
        \includegraphics[width=\textwidth]{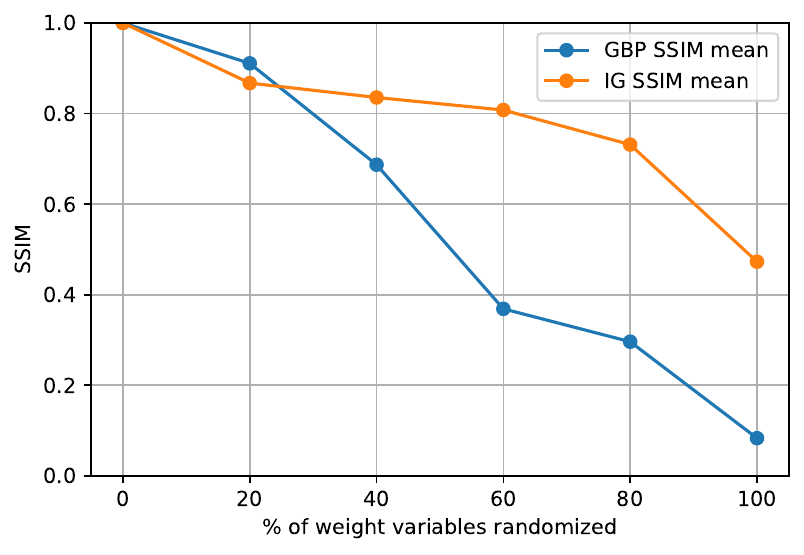}
        \caption{SSIM for the mean $\text{expl}_\mu(x)$}
        \label{weight_randomization_ssim_comparison_mean}
    \end{subfigure}
    \hfill
    \begin{subfigure}[b]{0.475\textwidth}
        \centering
        \includegraphics[width=\textwidth, ]{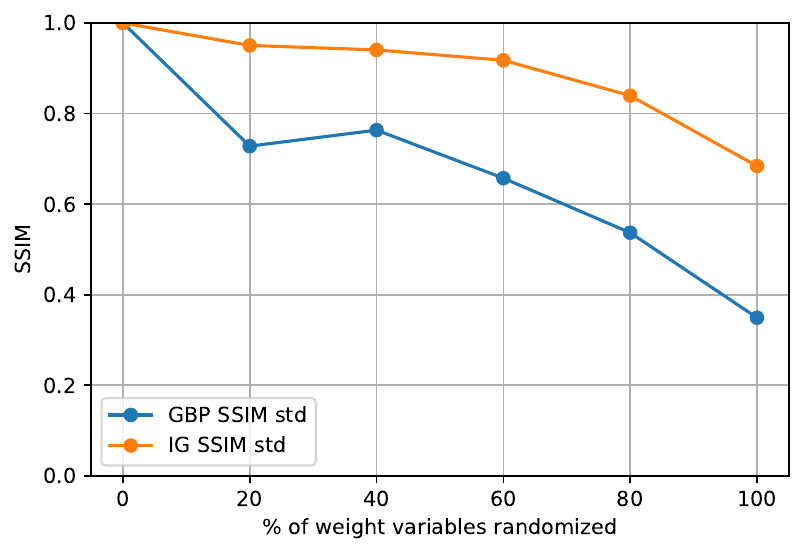}
        \caption{SSIM for the std $\text{expl}_\sigma(x)$}
        \label{weight_randomization_ssim_comparison_std}
    \end{subfigure}
    \caption{SSIM values for the weight randomization test on CIFAR 10 using Dropout. These values are computed between the explanation representation generated with no weight randomization and with incremental weight randomization. As the amount of weight randomization increases, the similarity between explanation representations decreases.}
    \label{weight_randomization_ssim_comparison} 
\end{figure}

\textbf{Weight randomization test}. Our main results using SSIM are shown in Figure \ref{weight_randomization_ssim_comparison}, where we compare the SSIM for $\text{expl}_\mu(x)$ and $\text{expl}_\sigma(x)$ separately. The mean behaves as expected, decreasing as more weights are randomized in both explanation methods, but GBP decreases to a minimum SSIM of around 0.1. The standard deviation $\text{expl}_\sigma(x)$ also decreases as expected in both explanation methods, but does not decrease at the same level as the mean, and its not close to being a minimum. We believe these results show that Dropout + GBP and Dropout + IG both pass the weight randomization test.

\begin{figure}[p]
    \begin{tabular}{p{1.3cm}l}
        \rotatebox{90}{Input} & \includegraphics[width=0.11\textwidth]{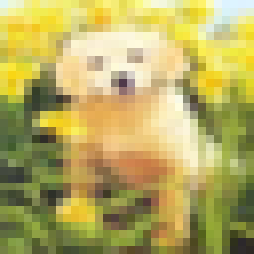}\\
    \end{tabular}
    
    \begin{tabular}{p{0.11\textwidth}p{0.14\textwidth}p{0.14\textwidth}p{0.14\textwidth}p{0.14\textwidth}p{0.15\textwidth}p{0.14\textwidth}}
        \tiny
        & 0\% & 20\% & 40\% & 60\% & 80\% & 100\%
    \end{tabular}
    
    \begin{minipage}{0.09\textwidth}
        \begin{tabular}{l}
            \vspace*{0.3cm} \rotatebox{90}{Mean} \vspace*{0.4cm}\\  \rotatebox{90}{Std}\vspace*{0.5cm}\\ \rotatebox{90}{Coef. Var.}
        \end{tabular}
    \end{minipage}
    \begin{minipage}{0.90\textwidth}
        \includegraphics[width=\textwidth]{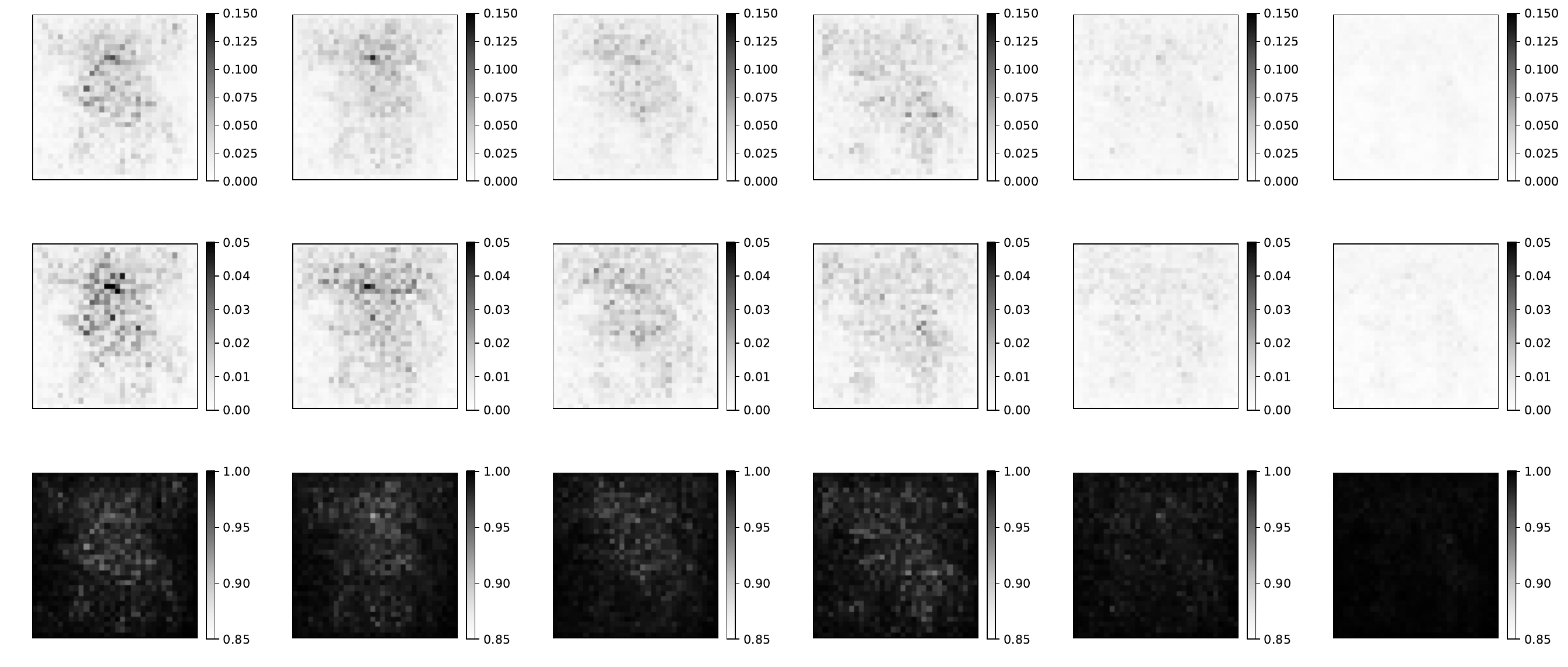}
    \end{minipage}
    
    \caption{Visualization of weight randomization effect on the explanations obtained using Dropout and Integrated Gradients on a CIFAR10 sample. Note how the both mean and standard deviation explanation become almost blank with increasing random weights.}
    \label{weight_randomization_dropout_ig}
\end{figure}

\begin{figure}[p]
    \begin{tabular}{p{1.3cm}l}
        \rotatebox{90}{Input} & \includegraphics[width=0.11\textwidth]{results/cifar-dog.png}\\
    \end{tabular}
    
    \begin{tabular}{p{0.11\textwidth}p{0.14\textwidth}p{0.14\textwidth}p{0.14\textwidth}p{0.14\textwidth}p{0.15\textwidth}p{0.14\textwidth}}
        \tiny
        & 0\% & 20\% & 40\% & 60\% & 80\% & 100\%
    \end{tabular}
    
    \begin{minipage}{0.09\textwidth}
        \begin{tabular}{l}
            \vspace*{0.3cm} \rotatebox{90}{Mean} \vspace*{0.4cm}\\  \rotatebox{90}{Std}\vspace*{0.5cm}\\ \rotatebox{90}{Coef. Var.}
        \end{tabular}
    \end{minipage}
    \begin{minipage}{0.90\textwidth}
        \includegraphics[width=\textwidth]{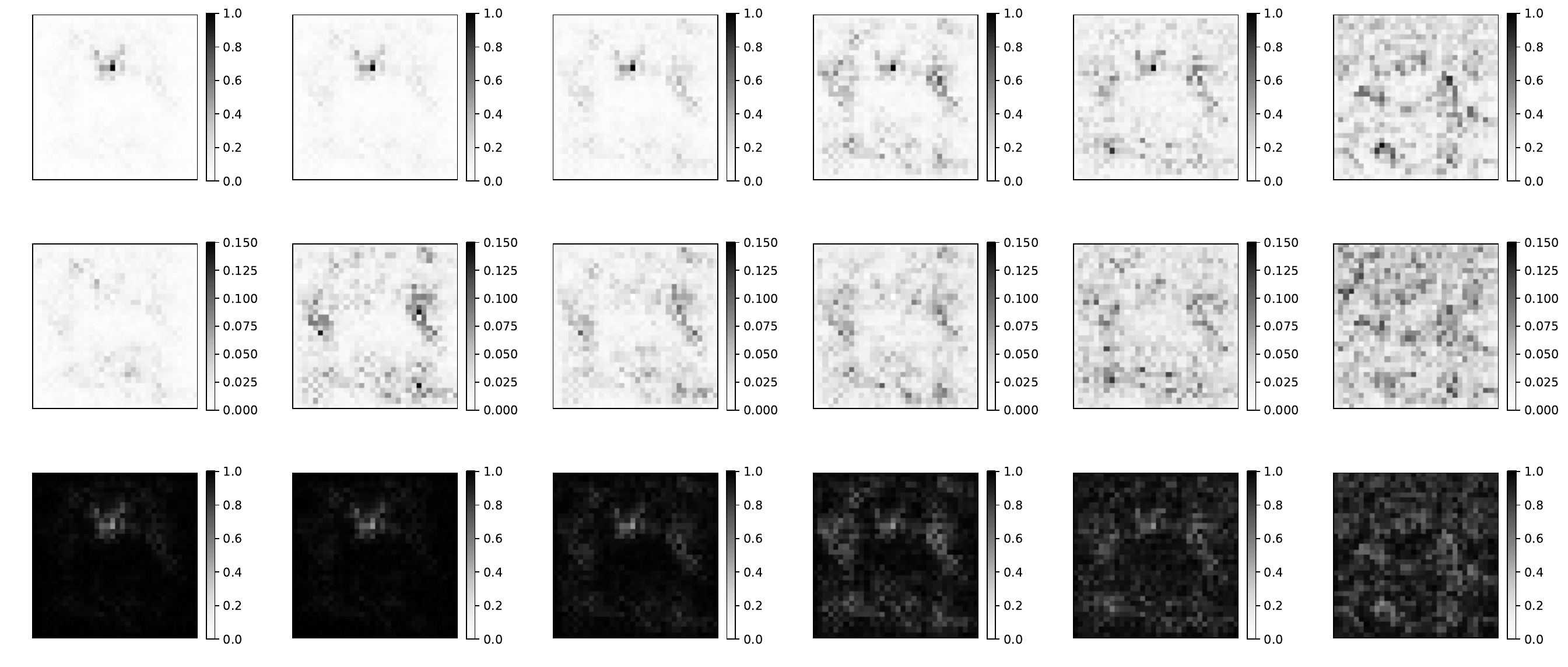}
    \end{minipage}
            
    \caption{Visualization of weight randomization effect on the explanations obtained using Dropout and Guided Backpropagation on a CIFAR10 sample. Note how the mean and standard deviation explanation become noisy with increasing random weights.}
    \label{weight_randomization_dropout_gbp}
\end{figure}

Figures \ref{weight_randomization_dropout_ig} and \ref{weight_randomization_dropout_gbp} show an individual sample and how their explanation mean, standard deviation, and coefficient of variation, vary with weight randomization. These particular examples show how these method pass the weight randomization test, as the mean and standard deviation explanations become more noisy with more layers in the network being randomized (as expected), with varying level of noise. It is clear how GBP explanations (Figure \ref{weight_randomization_dropout_gbp}) become more noisy with more random weights, while IG (Figure \ref{weight_randomization_dropout_ig}) produces an almost blank image, which explains the behavior presented in the SSIM results. Additionally these results show how differently GBP and IG work, producing very different behavior on the explanation and its uncertainty $\text{expl}_\sigma(x)$.

\FloatBarrier

\textbf{Data randomization test}. Main results for this test are shown in Figure \ref{data_randomization_ssim}, where we present SSIM values for $\text{expl}_\mu(x)$ and $\text{expl}_\sigma(x)$ separately. These values should be compared to a theoretical SSIM maximum value of 1.0 when no randomization is applied, and when randomizing labels, we obtain similar results compared to the weight randomization tests, where GBP decreases SSIM considerably more than IG, but overall both methods pass the data randomization test.

\begin{figure}[tb!]
    \begin{subfigure}{\textwidth}
        \centering
        \includegraphics[width=0.11\textwidth]{results/cifar-dog.png}
        \caption{Input Image}
    \end{subfigure}
    
    \begin{subfigure}{0.49\textwidth}
        \begin{tabular}{p{0.11\textwidth}p{0.42\textwidth}p{0.42\textwidth}}
            & True Labels & Random Labels
        \end{tabular}
        
        \begin{minipage}{0.09\textwidth}
            \begin{tabular}{l}
                \vspace*{0.3cm} \rotatebox{90}{Mean} \vspace*{1.2cm}\\  \rotatebox{90}{Std}\vspace*{1.2cm}\\ \rotatebox{90}{Coef. Var.}
            \end{tabular}
        \end{minipage}
        \begin{minipage}{0.90\textwidth}
            \includegraphics[width=\textwidth]{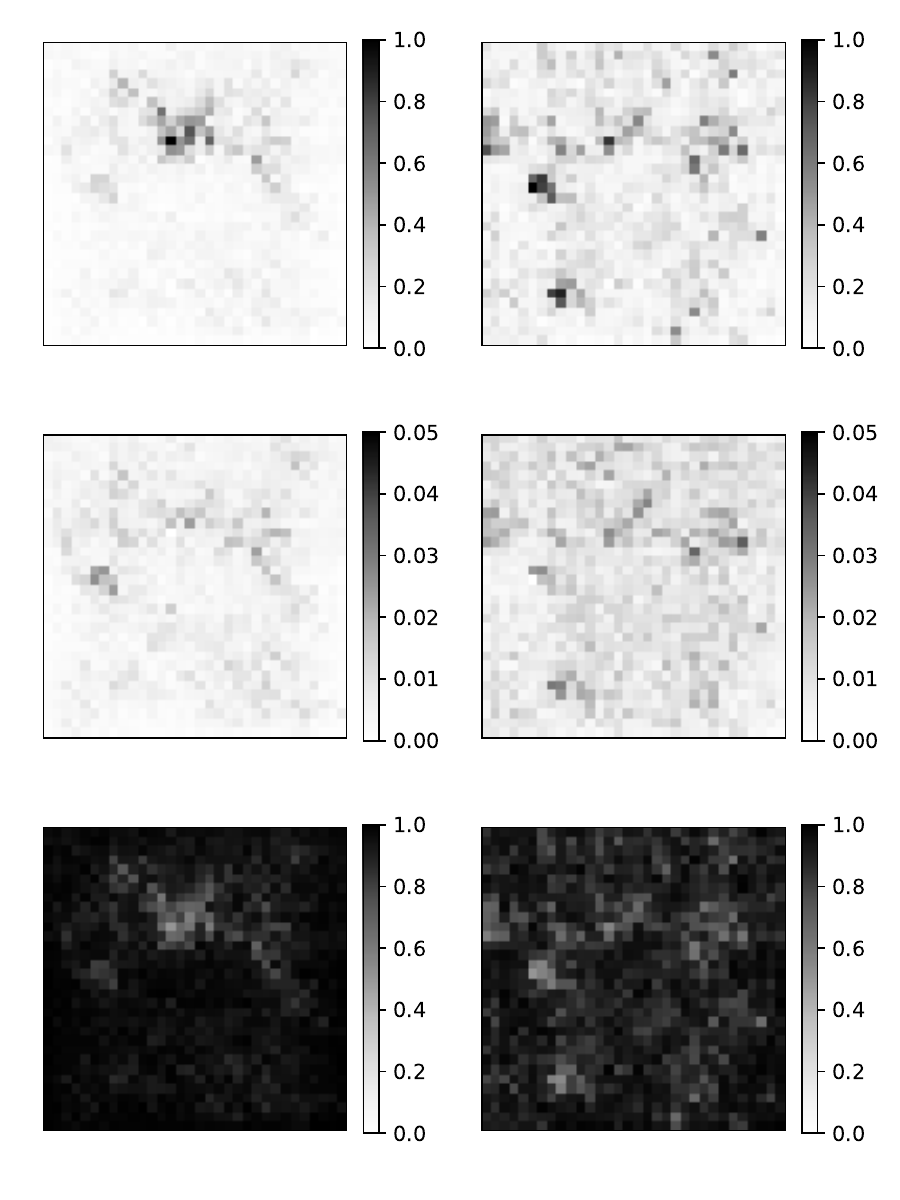}
        \end{minipage}
        
        \caption{Integrated Gradients}
        \label{data_randomization_ig}
    \end{subfigure}
    \begin{subfigure}{0.49\textwidth}
        \begin{tabular}{p{0.03\textwidth}p{0.42\textwidth}p{0.42\textwidth}}
            & True Labels & Random Labels
        \end{tabular}
        \includegraphics[width=0.9\textwidth]{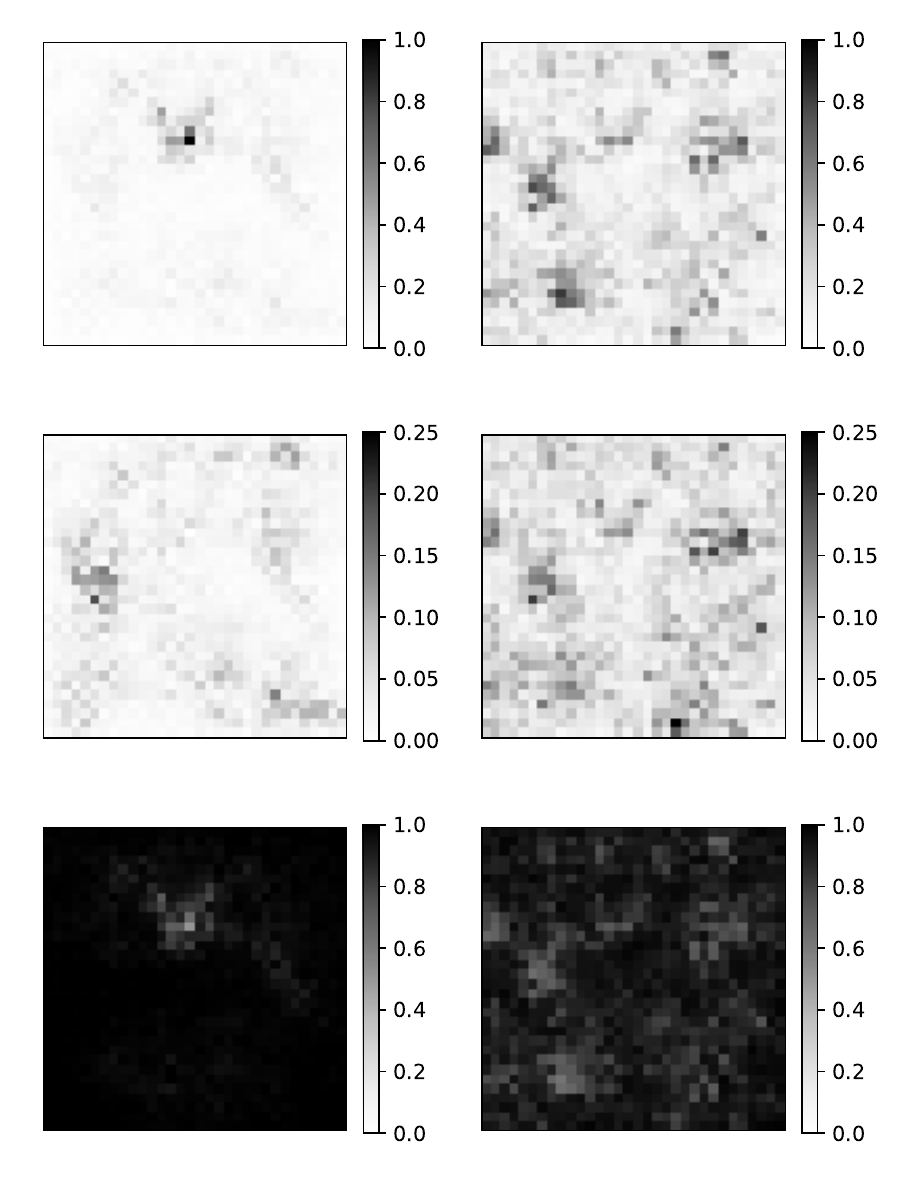}
        \caption{Guided Backpropagation}
        \label{data_randomization_gbp}
    \end{subfigure}    
    \caption{Visualization of data randomization effect on explanations obtained using Dropout and GBP and IG. Note how the explanation for the model with random labels is considerably more noisy than explanation from a model with true labels.}     
    \label{data_randomization_sample_comparison}
\end{figure}

Figure \ref{data_randomization_sample_comparison} shows one individual sample and the predicted explanations and their uncertainties for each stage of the data randomization test (model trained on true labels vs random labels). It is clear that both examples pass the data randomization test, as in both cases training on random labels produces a more noisy mean and standard deviation explanation, which is reflected into a noisy coefficient of variation. But it can be visually seen that GBP produces higher $\text{expl}_\sigma(x)$ values, indicating its further decrease of SSIM over the whole dataset.

\begin{table}[t]
    \centering
    \begin{tabular}
        {p{3cm}
            p{2cm} %
            p{2cm} %
        }
        \toprule
        \multicolumn{1}{p{3cm}}{\textbf{Explanation Method}} &  
        \multicolumn{1}{p{2cm}}{\textbf{Mean}} &  
        \multicolumn{1}{p{2cm}}{\textbf{Standard Deviation}} \\
        & $\text{expl}_\mu(x)$ & $\text{expl}_\sigma(x)$\\
        \midrule
        IG &  0.401 & 0.894
        \\
        GBP & 0.252 & 0.510
        \\
        \bottomrule
    \end{tabular}
    \caption{SSIM values for the data randomization on CIFAR10 using MC-Dropout. These values are computed between the explanation representation with no data randomization and with data randomization. These SSIM values should be compared to a theoretical maximum of 1.0 when no data randomization is performed. Note how GBP has a larger drop compared to IG.}
    \label{data_randomization_ssim}
\end{table}

\subsection{California Housing Regression}

In this section we evaluate a tabular regression dataset: the california housing dataset \cite{pace1997sparse}. We train a multi-layer perceptron with four fully connected layers, 8-8-8-1 neurons, and ReLU activation. Uncertainty estimation methods are applied to the last layer only, when applicable.

As tabular features are not directly humanly interpretable as images are, we compare the explanation uncertainty $\text{expl}_\sigma(x)$ directly for each test, using bar plots.

\begin{figure}
    \begin{tabular}{p{0.2cm}cc}
        & \textbf{Guided Backprop} & \textbf{LIME}\\
        \rotatebox{90}{\hspace*{1.0cm}\textbf{Dropout}} & \includegraphics[width=0.46\textwidth]{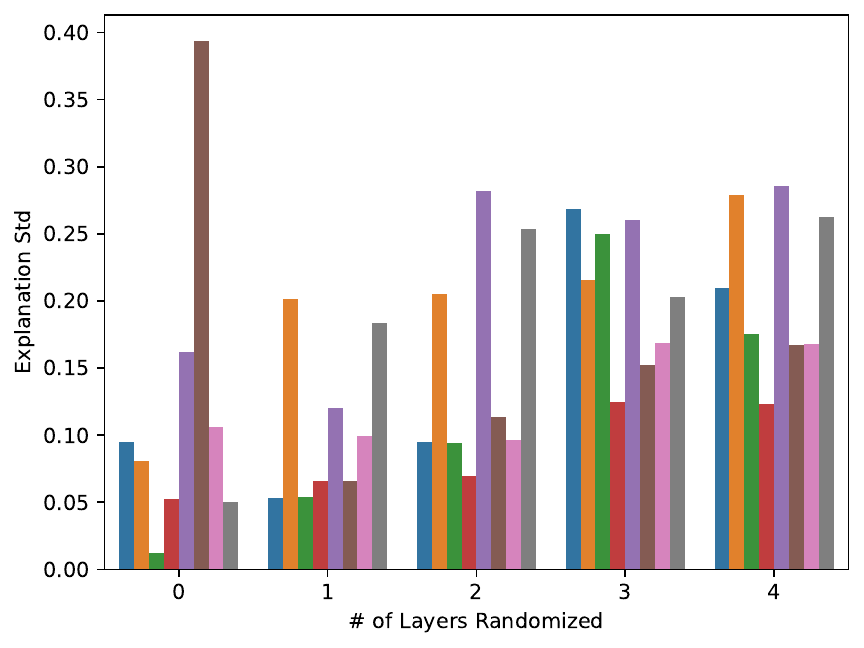} & \includegraphics[width=0.46\textwidth]{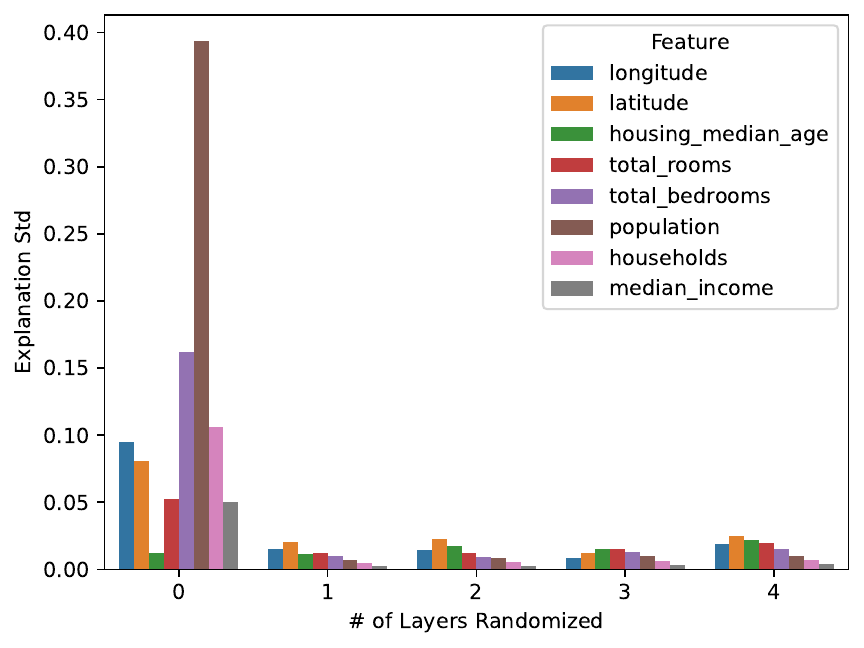}\\
        
        \rotatebox{90}{\hspace*{1.0cm} \textbf{DropConnect}} & \includegraphics[width=0.46\textwidth]{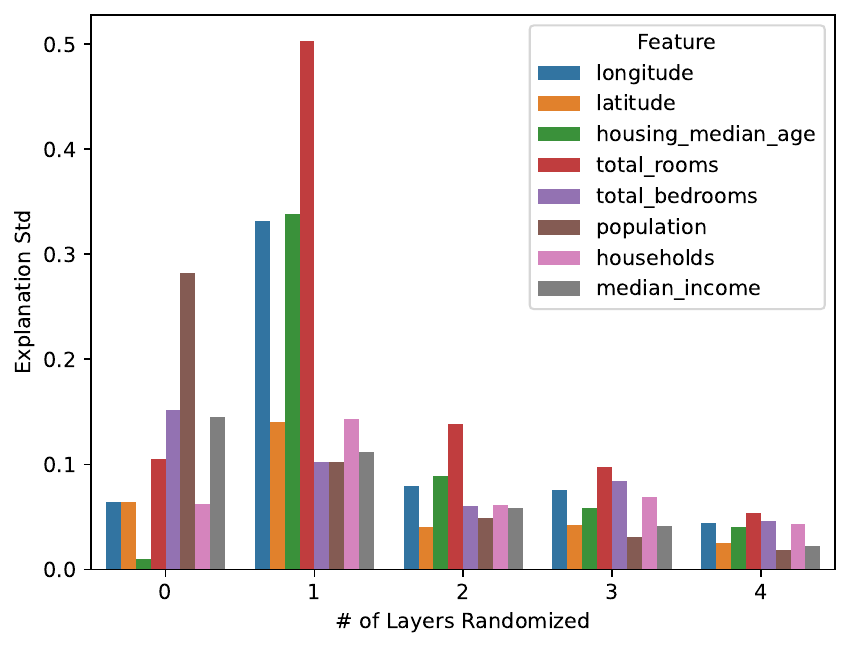} & \includegraphics[width=0.46\textwidth]{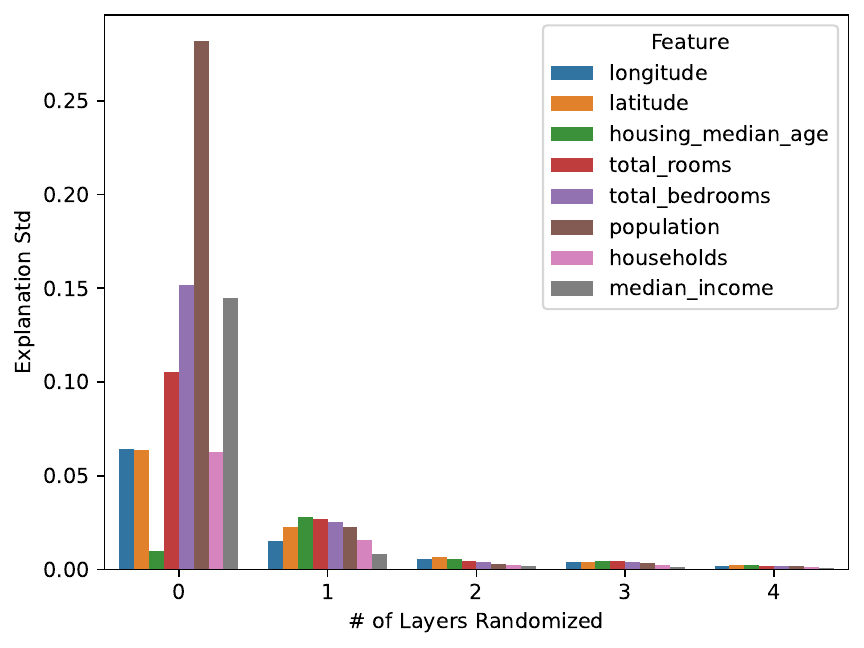}\\
        
        \rotatebox{90}{\hspace*{1.0cm} \textbf{Flipout}} & \includegraphics[width=0.46\textwidth]{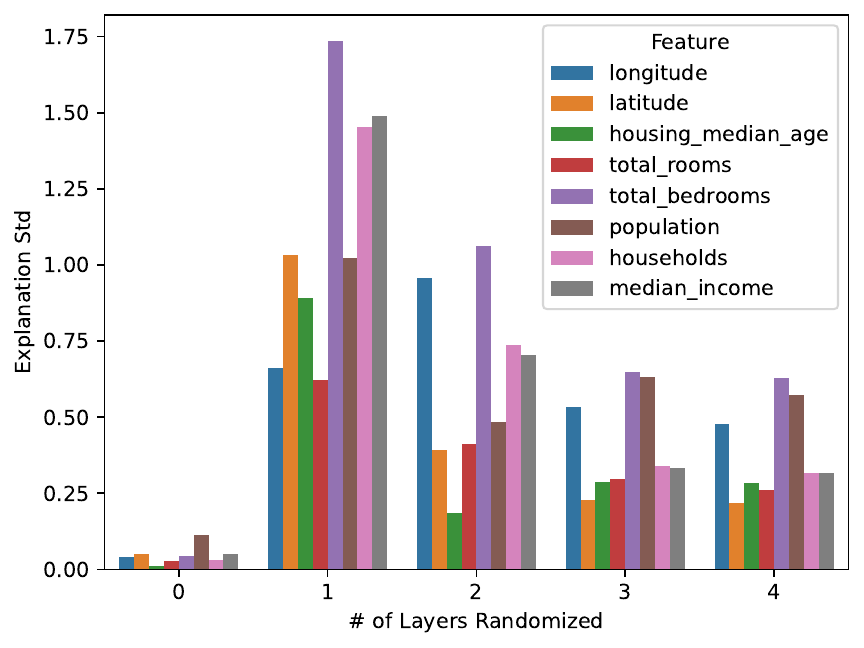} & \includegraphics[width=0.46\textwidth]{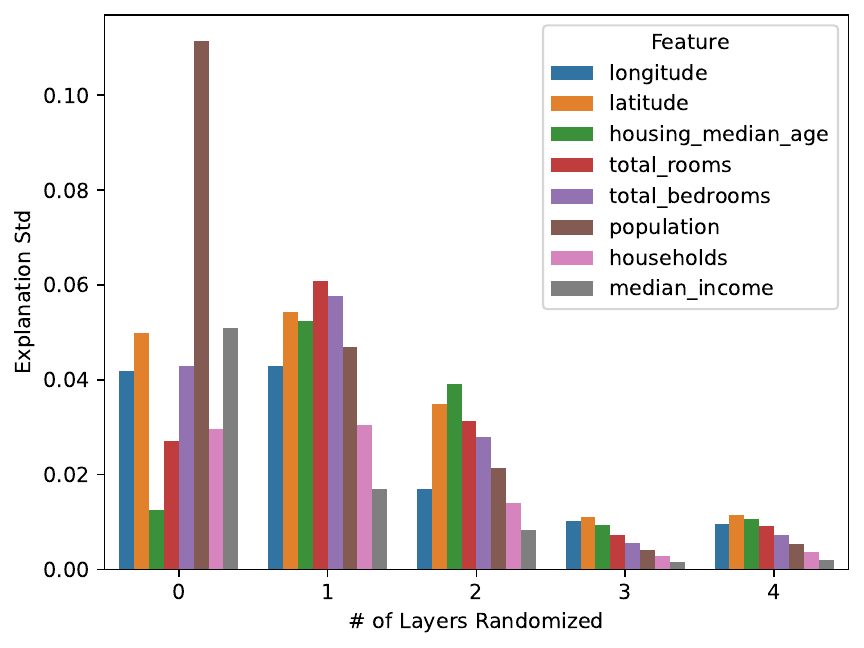}\\
        
        \rotatebox{90}{\hspace*{1.0cm} \textbf{Ensembles}} & \includegraphics[width=0.46\textwidth]{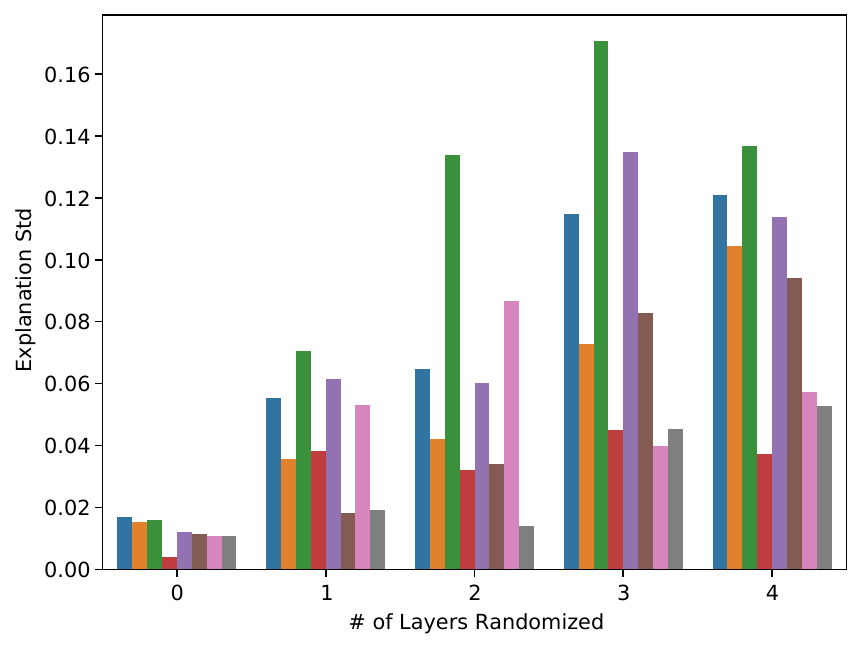} & \includegraphics[width=0.46\textwidth]{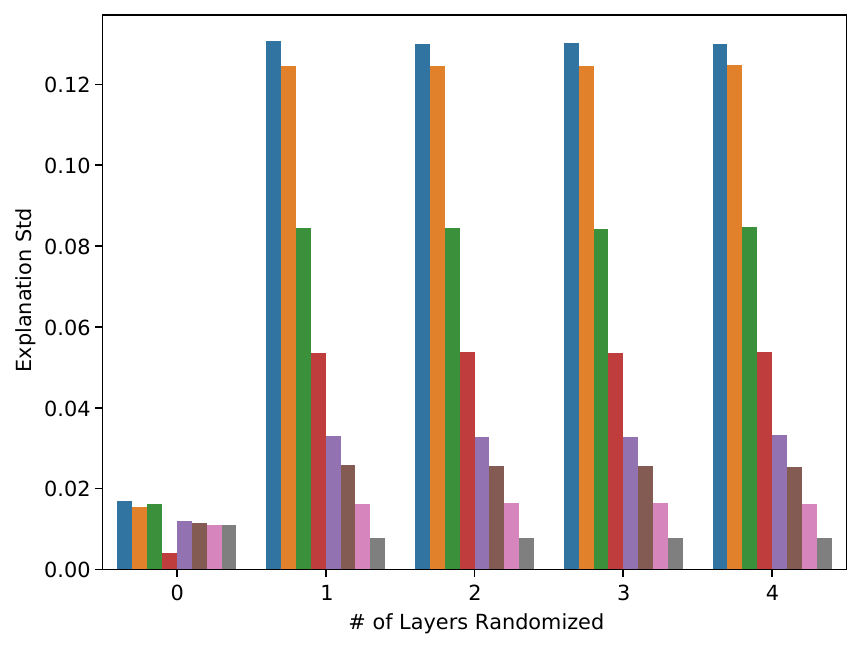}
    \end{tabular}
    \caption{Weight randomization test comparison across two explanation methods and four uncertainty methods on the California Housing Dataset (Tabular Regression). Dropout + GBP and Ensembles with GBP and LIME pass the test, while other combinations fail the test.}
    \label{weight_randomization_calif}
\end{figure}

\textbf{Weight randomization test}.  Figure \ref{weight_randomization_calif} presents the weight randomization results. Overall only ensembles consistently passes this test by producing higher explanation uncertainty for increasing randomized layer weights for both GBP and LIME. In the case of LIME, ensembles produced a constant higher explanation uncertainty, and we consider this to be passing the test.

Other uncertainty estimation methods produce inconsistent results with both GBP and LIME, like Flipout consistently producing lower explanation uncertainty as layers are randomized, which is counterintuitive to proposed interpretation of the weight randomization test.

\begin{figure}
    \begin{tabular}{p{0.2cm}cc}
        & \textbf{Guided Backprop} & \textbf{LIME}\\
        \rotatebox{90}{\hspace*{1.0cm}\textbf{Dropout}} & \includegraphics[width=0.46\textwidth]{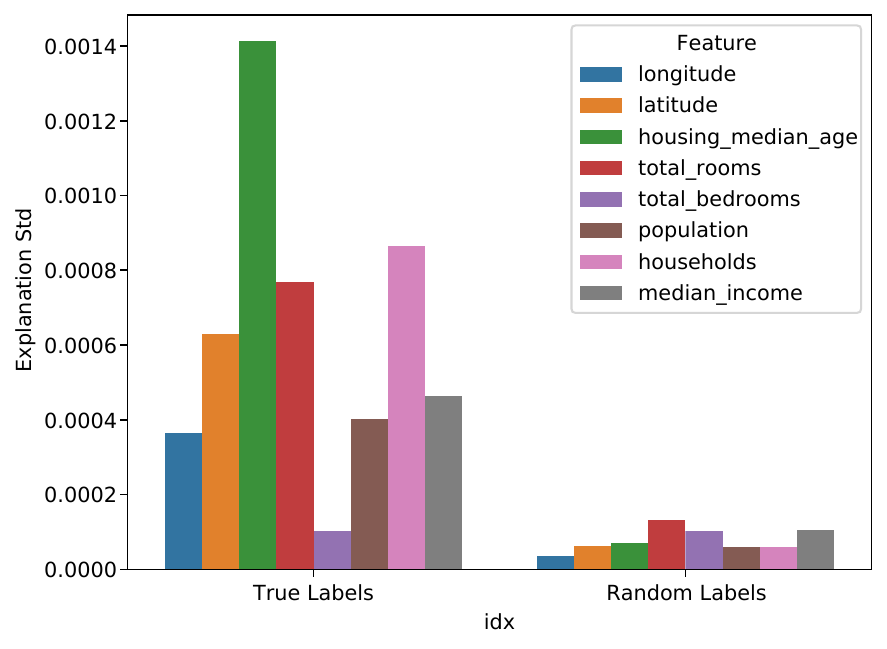} & \includegraphics[width=0.46\textwidth]{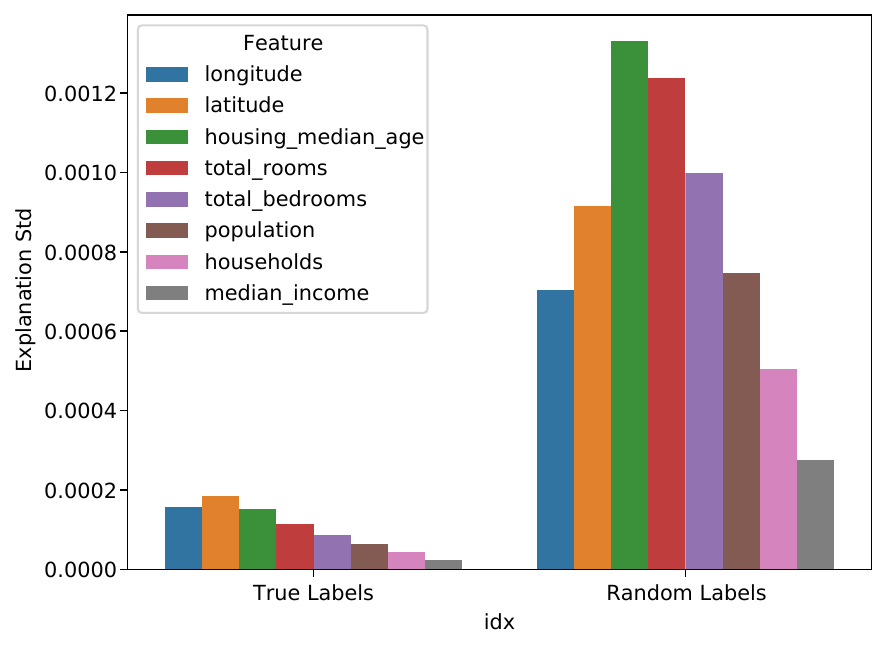}\\
        
        \rotatebox{90}{\hspace*{1.0cm} \textbf{DropConnect}} & \includegraphics[width=0.46\textwidth]{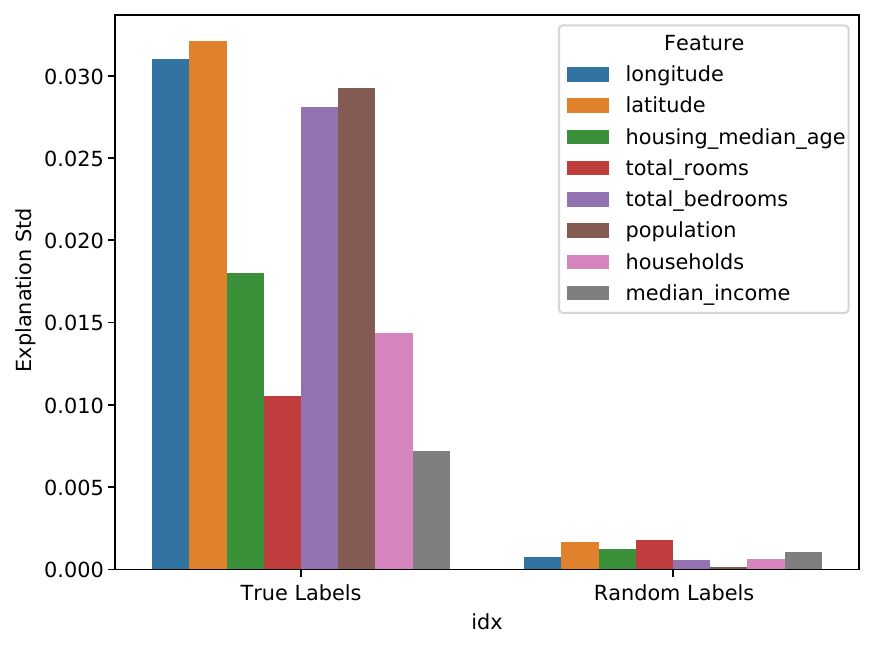} & \includegraphics[width=0.46\textwidth]{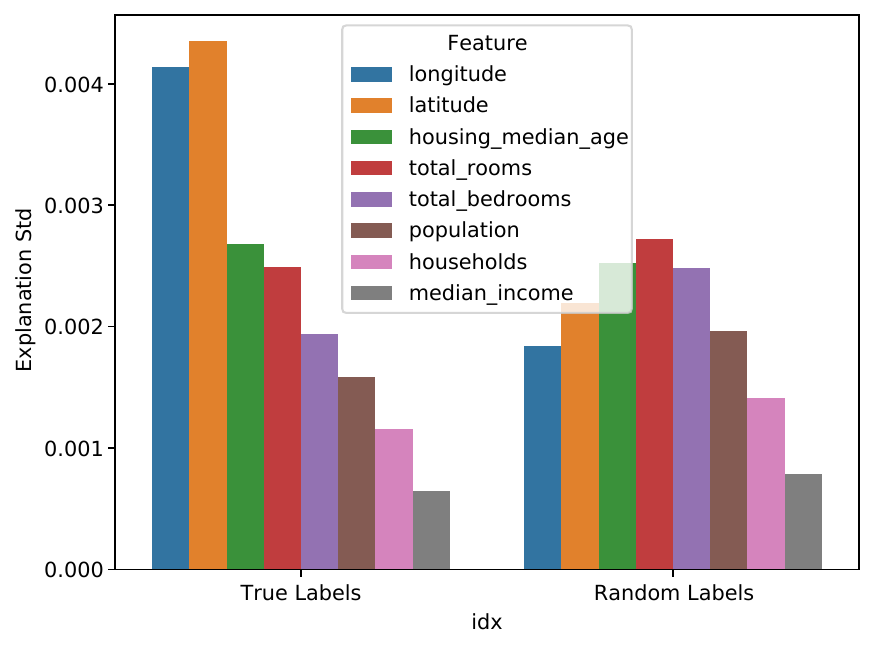}\\
        
        \rotatebox{90}{\hspace*{1.0cm} \textbf{Flipout}} & \includegraphics[width=0.46\textwidth]{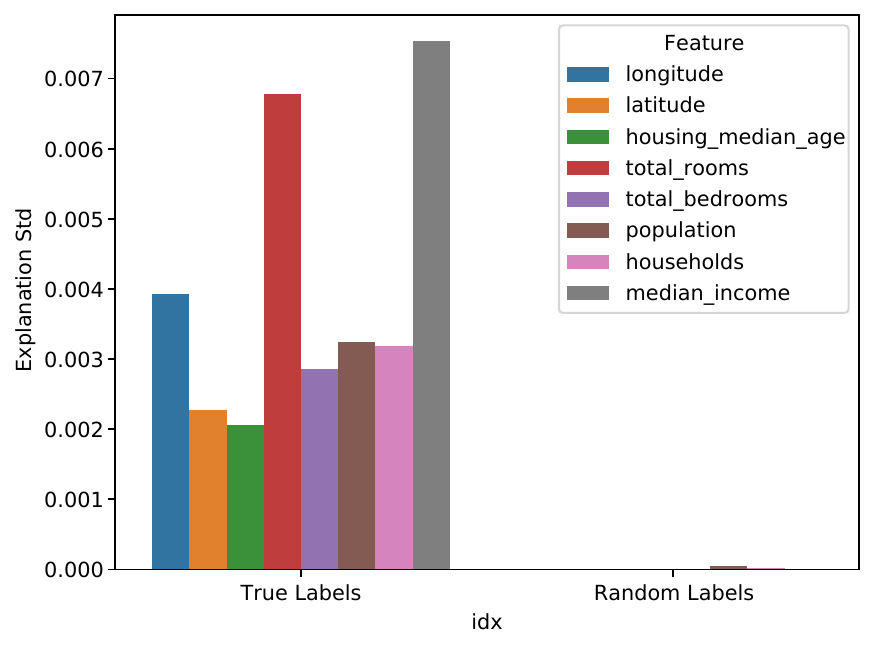} & \includegraphics[width=0.46\textwidth]{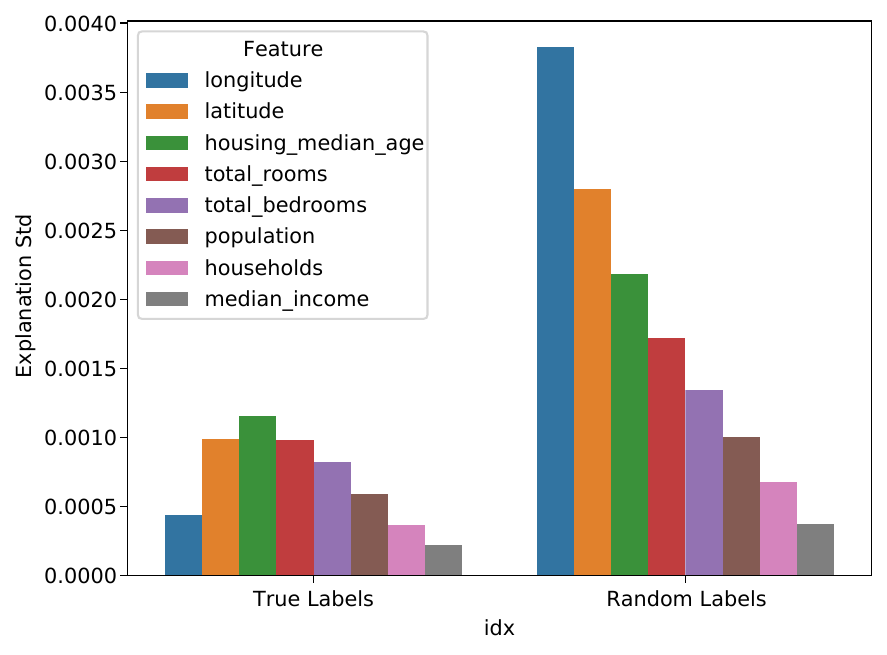}\\
        
        \rotatebox{90}{\hspace*{1.0cm} \textbf{Ensembles}} & \includegraphics[width=0.46\textwidth]{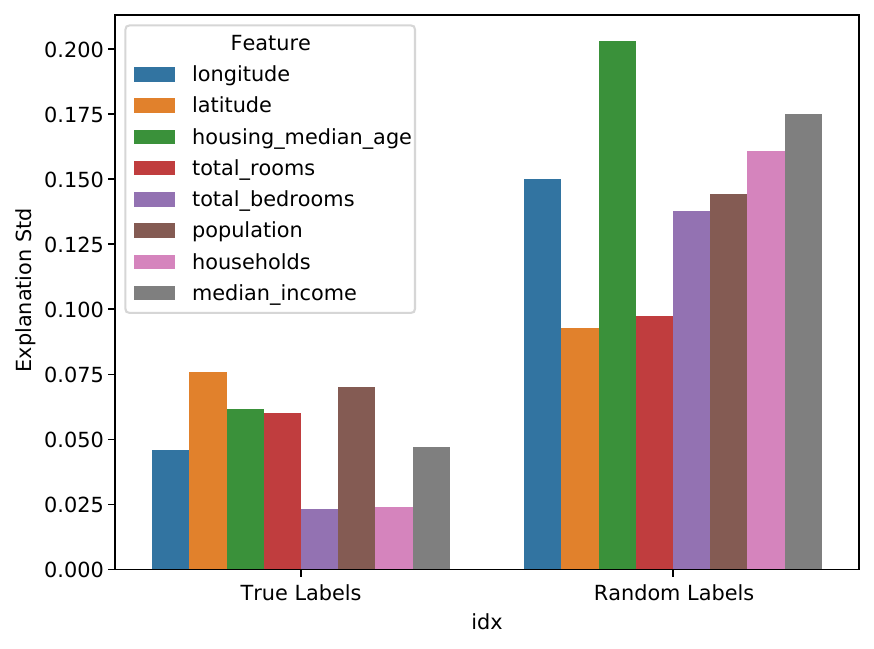} & \includegraphics[width=0.46\textwidth]{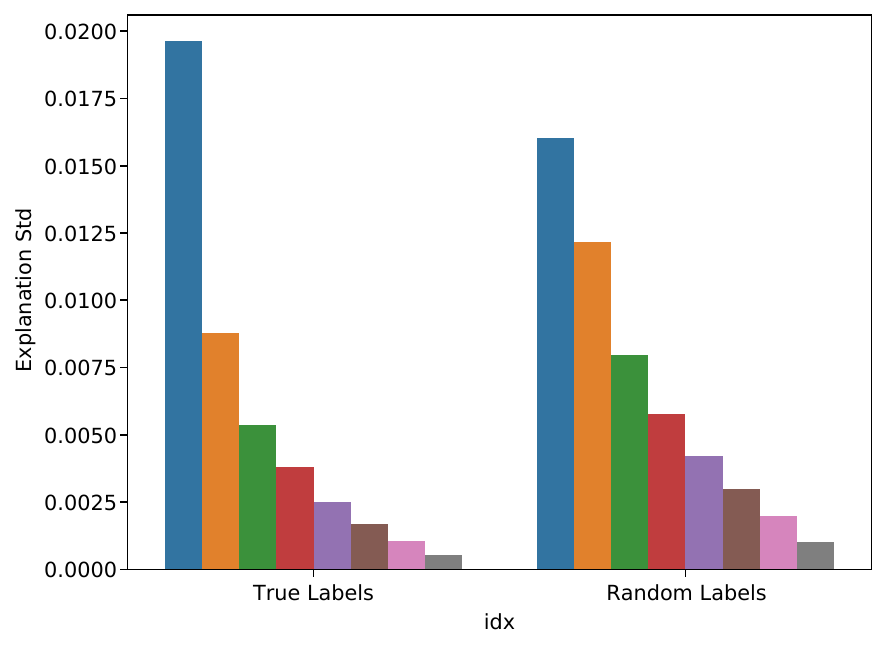}
    \end{tabular}
    \caption{Data randomization test comparison across two explanation methods and four uncertainty methods on the California Housing Dataset (Tabular Regression). Most combinations fail this test, with only Dropout + LIME, Flipout + LIME, and Ensembles with GBP and LIME passing the test.}
    \label{data_randomization_calif}
\end{figure}

\textbf{Data randomization test}. Figure \ref{data_randomization_calif} presents data randomization results. In this case we only compare the explanation uncertainty for a model trained with the true labels in the training set, versus another model trained with permuted labels, breaking the relationship between inputs and labels, and producing an overfitted model that does not generalize to the validation/test set.

Results are overall similar than with the weight randomization test, with ensembles clearly producing higher explanation uncertainty for random labels for both GBP and LIME, while other methods like Flipout and Dropout only sometimes produce this behavior, depending on the explanation method being used.

\subsection{Discussion}

Table \ref{sanity_checks_summary} provides as summary of all our results over multiple datasets, explanation methods, and uncertainty quantification methods. From our combined evaluation, it is clear that both MC-Dropout and Ensembles are the best uncertainty methods across all combinations of explanation method (GBP, IG, LIME), while other uncertainty estimation methods do not always pass both kinds of tests.

We believe that our results show that explanation methods and uncertainty quantification methods do have interacting effects, due to some UQ methods like ensembles producing consistent passing of both weight and data randomization tests, while other combinations do not consistently pass both tests, and only pass them in specific combinations with a explanation method.

Overall we believe that our evaluation clearly shows the usefulness of sanity checks for explanation uncertainty, which adds a useful tool to the explainable AI researchers' toolbox, in particular for explanation methods that can produce explanation uncertainty.

\begin{table}[tb!]
    \centering
    \begin{tabular}{lrccp{0.2cm}ccp{0.2cm}cc}
        \toprule
        &		   & \multicolumn{2}{c}{\textbf{GBP}} &  & \multicolumn{2}{c}{\textbf{IG}} & & \multicolumn{2}{c}{\textbf{LIME}}\\
        \midrule
        \textbf{Dataset} 	& \textbf{UQ Method} & Weight & Data & & Weight & Data & & Weight & Data\\
        \midrule
        \textbf{CIFAR10}					& Dropout & \cmark & \cmark &  & \cmark & \cmark  & &  & \\	 
        
        \midrule
       \textbf{Calif Housing}				& Dropout & \cmark & \xmark & & & & & \xmark & \cmark \\	 
        & DropConnect & \xmark & \xmark & & & & & \xmark & \xmark \\
        & Flipout & \xmark & \xmark &  &  & & & \xmark & \cmark \\                            
        & Ensembles & \cmark & \cmark & & & & & \cmark & \cmark \\        
        \bottomrule
    \end{tabular}    
    \caption{Summary of sanity checks for uncertainty explanations. A blank cell indicates that this combination of dataset/uncertainty method/explanation method was not tested.}
    \label{sanity_checks_summary}
\end{table}

\section{Conclusions and Future Work}

This work extends the common sanity checks for saliency explanations into explanation uncertainty methods, where a uncertainty estimation method is combined with an explanation method, to produce explanations with uncertainty, with aims to evaluate correctness of each explanations.

Our proposed weight and data randomization tests come with specific expectations, that increasing randomization in a model should be reflected on the explanation uncertainty $\text{expl}_\sigma(x)$.

We evaluate and showcase our tests on CIFAR10 and California Housing datasets, with multiple uncertainty estimation and explanation methods. Overall we find that Dropout and Ensembles with Guided Backpropagation, Integrated Gradients, or LIME, always pass our tests, while other combinations fail these tests.

We expect that our proposed tests for explanation uncertainty can be used to validate new developments in uncertainty estimation and saliency explaination methods.

Our work is limited by the selection of uncertainty estimation and saliency explanation methods, and by the selection of datasets used for evaluation. Additionally there could be other ways to compare explanation uncertainty in each test (we used SSIM and direct $\text{expl}_\sigma(x)$ comparisons), which needs to be adapted to specific use cases and input modalities.

\section*{Broader Impact Statement}

Explanations can be legally and ethically required, so users expect them to be correct or at least informative, but that not guaranteed with current explainable AI methods. We believe explanation uncertainty can have a positive societal impact, by quantifying uncertainty in explanations, and our proposed method can help evaluate and set basic expectations on explanations with uncertainty.

Uncertainty estimation is only approximate and can also produce misleading predictions, and when combined with explanation methods, it can produce misleading explanation uncertainty. There are no guarantees about correctness for both prediction, uncertainty, and explanation.

\clearpage
\bibliographystyle{splncs04}
\bibliography{mybibliography}

\end{document}